\documentclass{article}
\usepackage{spconf,amsmath,graphicx}

\usepackage{amsmath,amssymb,amsfonts}
\usepackage{algorithmic}
\usepackage{graphicx}
\usepackage{textcomp}
\usepackage{xcolor}
\usepackage{microtype}
\usepackage{hyperref}

\usepackage[all]{background}
\usepackage{url}
\usepackage{lipsum}

\usepackage{makecell}
\usepackage{multirow}
\usepackage{array}
\usepackage{booktabs}
\usepackage{enumitem}
\setlist{nosep, leftmargin=14pt}

\usepackage{mwe} 

\title{ConvLoRA and AdaBN based Domain Adaptation via Self-Training}

\name{Sidra Aleem\textsuperscript{1}, Julia Dietlmeier\textsuperscript{2}, Eric Arazo\textsuperscript{2}, Suzanne Little\textsuperscript{3} }
\address{\textsuperscript{1} SFI Research Centre for Machine Learning, Dublin City University, Ireland\\ \textsuperscript{2} Insight SFI Research Centre for Data Analytics, Dublin City University, Ireland\\ \textsuperscript{3} School of Computing, Dublin City University, Ireland}

\SetBgContents{Accepted for publication in IEEE ISBI 2024. Submission: 22 November 2023, Accepted: 2nd Feburary 2024. }
\SetBgPosition{current page.east}
\SetBgVshift{-0.5cm}
\SetBgOpacity{1.0}
\SetBgAngle{90.0}
\SetBgScale{-1.0}

\begin{document}
%
\maketitle
\begin{abstract}
Existing domain adaptation (DA) methods often involve pre-training on the source domain and fine-tuning on the target domain. For  multi-target domain adaptation, having a dedicated/separate fine-tuned network for each target domain,  that retain all the pre-trained model parameters, is prohibitively expensive. To address this limitation, we propose Convolutional Low-Rank Adaptation (ConvLoRA).
ConvLoRA freezes pre-trained model weights, adds trainable low-rank decomposition matrices to convolutional layers, and backpropagates the gradient through these matrices thus greatly reducing the number of trainable parameters. To further boost adaptation, we utilize Adaptive Batch Normalization (AdaBN) which computes target-specific running statistics and use it along with ConvLoRA. Our method has fewer trainable parameters and performs better or on-par with large independent fine-tuned networks (with less than 0.9\% trainable parameters of the total base model) when tested on the segmentation of Calgary-Campinas dataset containing brain MRI images. Our approach is simple, yet effective and can be applied to any deep learning-based architecture which uses convolutional and batch normalization layers. Code is available at: \href{https://github.com/aleemsidra/ConvLoRA}{ConvLoRA}.

\end{abstract}
\section{Introduction}
Deep neural networks (DNN) have achieved state-of-the-art performance when both train and test sets share the same distribution. However, domain shift, i.e. change in data distribution between train (source domain) and test (target domain) sets, significantly deteriorates the generalizability \cite{chen2018domain, ganin2016domain}. This issue is particularly pronounced in multi-center medical studies, where various imaging centers employ different scanners, protocols, and subject populations \cite{ganin2016domain, choudhary2020advancing}.


Unsupervised domain adaptation (UDA) \cite{chen2018domain, ganin2016domain} aims to generalize large-scale models, pre-trained on the source domain to an unlabeled target domain, eliminating the need for costly data annotation. It is typically achieved through fine-tuning, where a model pre-trained on the source domain is adapted to target domains. However, a major downside of fine-tuning is that it results in a dedicated model for each target domain with the same parameters as the original pre-trained model  \cite{hu2021lora,shirokikh2020first}. Consequently, several target domains would require several dedicated models with the same parameter count as the original pre-trained model. 

Thus UDA methods can be effective for single-target DA, resulting in a single model for a specific target domain. Conversely,  in multi-target DA (MTDA) the objective is to adapt to multiple unlabeled target domains. MTDA has a broader applicability to real-world scenarios. 
However, training separate models for each target domain with the same trainable parameters as the source model is impractical and prohibitively expensive. 

Parameter-efficient fine-tuning (PEFT) has demonstrated its effectiveness as a fine-tuning strategy for Large Language Models (LLMs) \cite{zaken2021bitfit}. Unlike conventional fine-tuning, it keeps the majority of the model parameters frozen while adapting a substantially reduced number of parameters, often less than 5\% of the total. This enables both efficient learning and faster updates. PEFT also outperforms full fine-tuning 
and enhances generalization, particularly in low-data scenarios  \cite{zaken2021bitfit}.

In the field of medical imaging, only a few methods have used adapter-based PEFT in Transformer-based architectures \cite{wang2023med, wu2023medical}. These works focus on achieving parameter-efficient transfer learning from natural images to medical images. To the best of our knowledge, both the application of PEFT in medical imaging in the context of UDA, and the use of adapter-based methods in CNNs have not yet been explored \cite{Hospidales_review2023}. 

Having identified this research gap, we propose a novel parameter-efficient MT UDA for medical image segmentation, that is computationally efficient and also has low-memory
footprint. \textbf{First}, we propose Convolutional Low-Rank Adaptation (ConvLoRA), as an adaptation of Low-Rank Domain Adaptation (LoRA) in LLMs \cite{hu2021lora}. ConvLoRA is specifically designed for application in Convolutional Neural Networks (CNNs), presenting a novel approach to address domain adaptation challenges in the context of image data. Instead of creating dedicated fine-tuned models for multiple target domains, each with the same number of parameters as the base model, we inject several ConvLoRA adapters into the base model pre-trained on the source domain, and only adapt the ConvLoRA parameters, while keeping all other parameters frozen. This method allows faster updates by adapting only a small set of domain specific parameters. 
\textbf{Second}, we further mitigate domain shift, introduced by statistical differences in mean and variance between source and target data, without requiring additional fine-tuning and computational resources. Instead of Batch Normalization (BN), we utilize Adaptive Batch Normalization (AdaBN) \cite{li2016revisiting},
which computes target-specific batch-wise running mean and variance, rather than using source domain's statistics.

\textbf{Our contributions can be summarized as follows}:
\begin{itemize}
    \item 

    Inspired by the recent advances in the LLMs, we propose a novel multi-target UDA approach that leverages the concept of our proposed parameter-efficient ConvLoRA adapter and AdaBN.
    To our best knowledge, this is the first work to adapt LoRA \cite{hu2021lora} to CNNs, particularly for UDA in the context of medical image segmentation.

    \item We show that our proposed UDA pipeline results in a significant reduction of over 99\% in trainable parameters while simultaneously achieving competitive segmentation accuracy compared to other methods.

    \item Our framework is generic, flexible and easily integrates with CNN-based architectures, significantly lowering training costs while enhancing adaptation.

\end{itemize}

\section{Related Work}
\textbf{Unsupervised Domain Adaptation (UDA)}
Several works employ adversarial learning, such as CycleGAN~\cite{zhu2017unpaired} and domain-invariant feature learning~\cite{hoffman2016fcns}, to adapt segmentation models~\cite{li2019bidirectional}. Huang et al.~\cite{huang2018domain} propose a method of matching layer-wise activations across domains.

\textbf{UDA for medical image segmentation}
An adversarial network is proposed for brain lesion segmentation in \cite{demner2016preparing}. Kushibar et al.~\cite{kushibar2019supervised} show that fine-tuning only the last CNN layer improves performance. However, it lacks a comparison with other DA methods. The last CNN layer is fine-tuned, but focus of this work is more on the training cases selection procedure rather than on adaptation~\cite{valindria2018domain}. 
Cross-modality DA for cardiac MR and CT image segmentation is achieved by adapting low-level layers \cite{zhuang2016multi}. Fine-tuning of early U-Net layers is done for skull segmentation~\cite{shirokikh2020first}.

\begin{figure}[!ht]
 \centering
    \includegraphics[width=0.85\columnwidth, height=6.4cm]{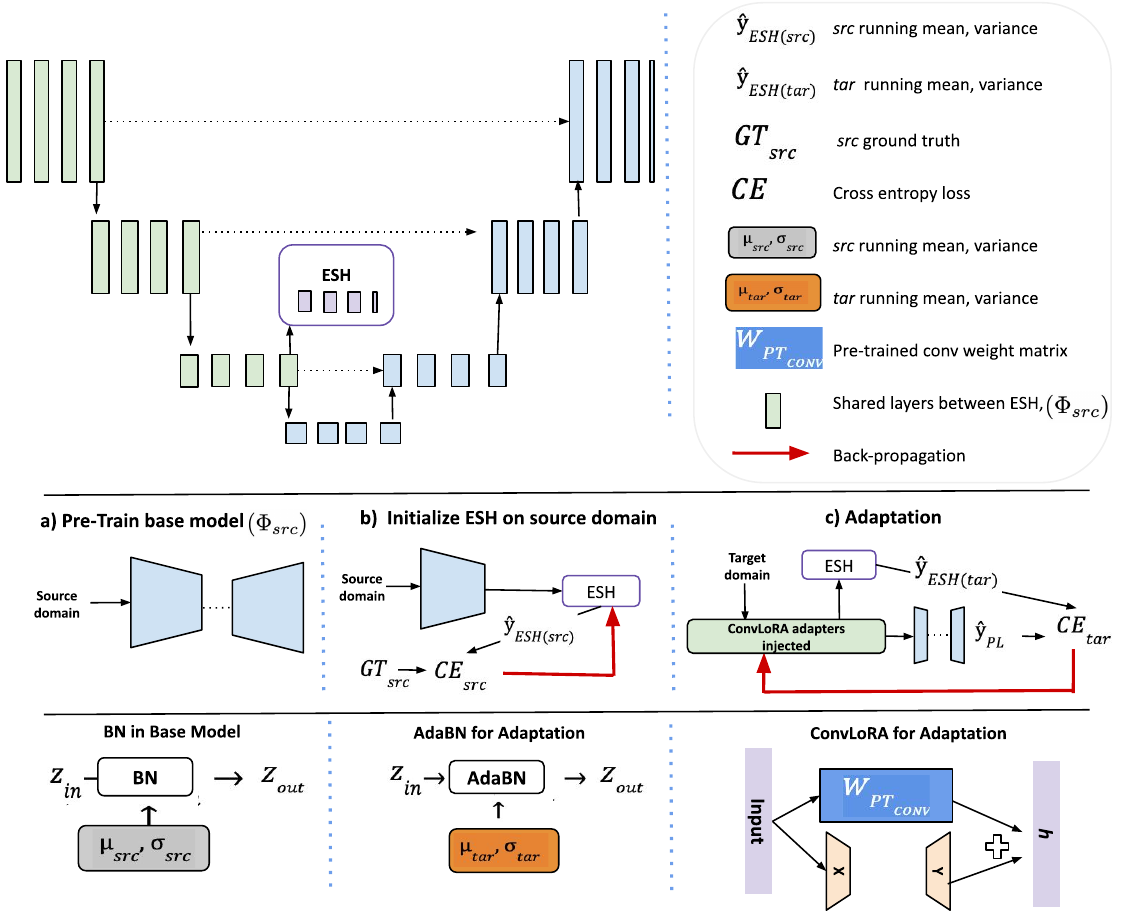}
    \caption{2D U-Net with Early Segmentation Head (ESH) is pre-trained on the source domain. ConvLoRA adapters facilitate adaptation in the encoder, along with AdaBN throughout the network.}
    \label{udas_arch}
\end{figure}

\textbf{Batch Normalization (BN)}
Chang et al.~\cite{chang2019domain} show that unsupervised fine-tuning of BN layers in the target domain improves adaptation. AdaBN,  computes mean and variance for BN running statistics in the target domain, enhances generalization \cite{li2016revisiting}. Test-time adaptation mitigates domain shift by recalculating running statistics for the current test input \cite{liu2021ttt++, choi2022improving, wang2022continual}.



\textbf{Parameter Efficient Fine Tuning (PEFT)} There are two prominent strategies for PEFT: a) adding adapter layers and only adapting them \cite{houlsby2019parameter}, b) optimizing some form of activations \cite{li2021prefix, lester2021power}. The use of adapters even in small networks leads to inference latency and extra compute \cite{hu2021lora}. LoRA minimizes latency by decomposing pre-trained weights into smaller matrices, fine-tunes only these matrices, and consequently lowers the memory usage \cite{hu2021lora}.

\begin{table*}[!ht]
\small
\centering
\caption{Surface Dice Score (SDS) on the CC359 \cite{souza2018open} dataset: comparison of different domain adaptation methods. Note that we are reporting mean and standard deviation results based on adaptation with three different seeds}

\begin{tabular}{lcccccc}
\toprule
\multirow{2}{*}{\textbf{\makecell{Target\\Domain}}}  & 
\multirow{2}{*}{\textbf{\makecell{Source\\Model}}} & 
\multirow{2}{*}{\textbf{\makecell{Self-\\Training \cite{zou2018unsupervised} }}} &

\multirow{2}{*}{\textbf{\makecell{UDAS\\\cite{sheikh2022unsupervised}}}} & 
\multirow{2}{*}{\textbf{\makecell{UDAS ConvLoRA\\ (Ours)}}} &
\multirow{2}{*}{\textbf{\makecell{ConvLoRA\\ + AdaBN (Ours)}}}  
\\
\\
\midrule
GE 1.5 & 0.734 ± 0.030 &  0.5304 & 0.7588 &  0.8368  ± 0.0386  &\textbf{0.8908 ± 0.0190} \\
Philips 1.5 & 0.871 ± 0.021 & 0.7252& 0.8460 & 0.8778 ± 0.0058 &\textbf{0.9143 ± 0.0121} \\
Philips 3 & 0.618 ± 0.005 & 0.6623 &   0.6623  & 0.7195  ± 0.0094&  \textbf{0.8251 ±  0.019 } \\
Siemens 1.5 & 0.825 ± 0.031 & 0.6929&  0.8245  & 0.8035  ± 0.0127 & \textbf{0.8923 ± 0.009}\\
Siemens 3 & 0.843 ± 0.012 & \textbf{ 0.8918 } & 0.8874 & 0.8494  ± 0.0026  & 0.8882 ± 0.006\\
\bottomrule
\end{tabular}

\label{tab_da:performance_stats}
\end{table*}
\section{Method}
Figure \ref{udas_arch}. provides an overview of our architecture. We integrate ConvLoRA and AdaBN into the UDAS model \cite{sheikh2022unsupervised} which consists of 2D U-Net with an added Early Segmentation Head (ESH). 
ESH consists of three convolutional layers, each followed by a BN layer. We inject ConvLoRA in the encoder part (see Figure. \ref{udas_arch}(c)) of the UDAS model and adapt it using the network's final predictions as pseudo-labels via self-training.

\subsection{ConvLoRA}
\label{convlora}
We propose a new ConvLoRA adapter, an extension of LoRA \cite{hu2021lora}, for parameter-efficient UDA for CNNs. 
For a pre-trained convolutional layer weight matrix $W_{PT_{CONV}} \in \mathbb{R}^{m\times n}$, ConvLoRA constrains its update by representing it with a low-rank decomposition: $W_{PT_{CONV}} + \Delta W_{CONV} = W_{PT_{CONV}} + XY$, where  $X \in \mathbb{R}^{m \times r} $ and $Y \in \mathbb{R}^{n \times r} $ are low-rank matrices and rank $r << min(m,n)$. During training, $W_{PT_{CONV}}$ is frozen, and does not receive gradient updates, while $X$ and $Y$ contain trainable parameters. Both $W_{PT_{CONV}}$ and $\Delta W_{CONV}$ are multiplied by the input and the respective output vectors 
are summed coordinate-wise. Hence, the forward pass operation is as follows:
\begin{equation}
    h = W_{PT_{CONV}} x +\Delta W_{CONV} x = W_{PT_{CONV}}x + XYx
\end{equation}
where $x$ is input, $X$ is initialized by random Gaussian distribution and $Y$ is zero in the beginning of training.

\subsection{Adaptive Batch Normalization (AdaBN)}
In this work, we use AdaBN \cite{li2016revisiting} instead of BN. While BN normalizes activation outputs using batch statistics, using source domain statistics for standardizing the target domain can lead to misclassification~\cite{mirza2022norm}. AdaBN computes the target domain-specific batch-wise mean and variance~\cite{li2016revisiting}. The standardization of each layer by respective domain ensures that each layer receives data from a similar distribution. 

\subsection{ConvLoRA and AdaBN based UDA}
\textbf{Baseline} Let $\Phi_{src}$ be the network trained solely with labeled source domain data $X_{src}$. Our goal is to adapt $\Phi_{src}$ to out-of-distribution unlabeled target data $Y_{tar}$ in a parameter-efficient unsupervised manner. As a backbone for $\Phi_{src}$, we use a U-Net architecture. We adopt the approach proposed in~\cite{sheikh2022unsupervised} as our baseline.

\textbf{Early Segmentation Head (ESH)} For the adaptation phase, a small CNN called ESH is placed after the encoder as shown in Figure~\ref{udas_arch}. We initialize ESH on the source domain by pre-training with the cross-entropy loss between the output of ESH and the ground truth mask. Then, during adaptation, target domain images are fed to both  $\Phi_{src}$ and ESH. The segmentation outputs from  $\Phi_{src}$ are used as pseudo-labels $(\hat{y}_{PL})$ to improve ESH predictions. At this stage, all the weights in $\Phi_{src}$ and ESH are frozen, except the encoder of $\Phi_{src}$ shared between the two networks. Since the encoder is shared between  $\Phi_{src}$ and ESH, improving ESH benefits $\Phi_{src}$.

\textbf{Adaptation}
In our proposed adaptation schema, all the parameters of the network $(\Phi_{src})$, other than the ConvLoRA parameters and running mean and running variance of BN layers are frozen. We integrate ConvLoRA adapter (discussed in Section~\ref{convlora}) into the encoder part of $\Phi_{src}$. While both $\Phi_{src}$ and ESH process the target domain images in the same manner as in~\cite{sheikh2022unsupervised}, we restrict gradient updates exclusively to the ConvLoRA adapter parameters. Consequently, we have a reduced number of domain-specific ConvLoRA parameters while having a single $\Phi_{src}$. To further mitigate the domain shift in a parameter-efficient way, we used the target domain's running mean and running variance, calculated via AdaBN. The source domain statistics are updated by computing target-specific batch-wise running statistics. Adapting the running mean and variance with AdaBN is straightforward and facilitates parameter-free adaptation without extra parameters and components, as these statistics are not trainable parameters.

\section{Experimental Setup}
We evaluate our approach on Calgary-Campinas (CC359) dataset \cite{souza2018open}, a multi-vendor (GE, Philips, Siemens), multi-field strength (1.5, 3) magnetic resonance (MR) T1-weighted volumetric brain imaging dataset. It has six different domains and contains 359 3D brain MR image volumes, primarily focused on the task of skull stripping. The source model $(\Phi_{src})$ is pre-trained on the GE 3 (source domain) using an 80:10:10 split. For adaptation, only 10 images from each target domain's training set are randomly chosen, and inference is conducted on the respective official test sets. Pre-processing involves removing black slices and min-max scaling, with all images resized to 256$\times$256 resolution.

The source model ($\Phi_{src}$) is trained for 100 epochs using a batch size of 32, a learning rate of 0.001, and optimized with the Adam optimizer using cross-entropy loss. The ESH is trained for 20 epochs, followed by our adaptation method which is trained for only 5 epochs with a learning rate of 0.0001. When using ConvLoRA, we set the rank to $r=2$, given that the original kernel weight was 3. Surface Dice Score (SDS)~\cite{shirokikh2020first} is used to assess the image segmentation performance. This metric is more informative than volumetric Dice as it  emphasizes on the brain contour over internal volume~\cite{shirokikh2020first} and it is widely used in methods exploring CC359 \cite{sheikh2022unsupervised, shirokikh2020first, zhuang2016multi}.

The processing pipeline was implemented in Python 3.8.17, and open-source library PyTorch 2.0.1 is used. All experiments were performed on a desktop computer with the Ubuntu operating system 20.04.6 LTS with CUDA 11.6, NVIDIA GeForce RTX 3090 GPU, and a total of 62 GB RAM.

\begin{table*}[!ht]

   \caption{Ablation Study: Placement of ConvLoRA adapters and respective SDS, (Enc: Encoder).}
\centering
\begin{tabular}{lccccc} 
    \toprule
    \multirow{2}{*}{\textbf{\makecell{Target\\Domain}}}  & 
    \multirow{2}{*}{\textbf{\makecell{Enc. \\ Block 1}}} & 
    \multirow{2}{*}{\textbf{\makecell{Enc. \\Block 1-2}}} &
    \multirow{2}{*}{\textbf{\makecell{Enc. \\ Block 1-3}}} &
    \multirow{2}{*}{\textbf{\makecell{Full Enc. \\Block }}} &
    \multirow{2}{*}{\textbf{\makecell{Full Enc. Block  +\\AdaBN}}} \\\\
    
    \midrule
        GE 1.5 &  0.8368 ± 0.0386 & 0.8275  ± 0.0118 &  0.8081 ± 0.0103 & 0.8611 ±  0.044
        & \textbf{0.8908 ± 0.019} \\
        Philips 1.5 &  0.8778 ± 0.0058 & 0.8329 ±  0.1029& 0.84046 ± 0.0380 & 0.8910 ± 0.0270 &\textbf{ 0.9023 ±  0.010} \\
        Philips 3 & 0.7195 ± 0.0094 & 0.7388 ± 0.0223  &  0.74979 ± 0.0146 & 0.7653 ± 0.0060 & \textbf{0.8251 ±  0.019 }\\
        Siemens 1.5 & 0.7195 ± 0.0094 & 0.8521 ± 0.0094 & 0.8610  ± 0.0284 & 0.8404 ± 0.0380 & \textbf{0.8923 ± 0.009} \\
        Siemens 3 & 0.8494  ± 0.0020 & 0.8560 ± 0.0171  & 0.8685  ± 0.0218 &  0.8584  ±  0.0139 & \textbf{0.8882 ± 0.006}

\\
    \bottomrule
    \end{tabular}
 
\label{tab:ablations}
\end{table*}
\textbf{Source Model} refers to the base model ($\Phi_{src}$) trained exclusively on the source data, without any adaptation to target domains. \textbf{Self-Training} employs pseudo-labels of the target domain to iteratively enhance model performance \cite{zou2018unsupervised}. \textbf{UDAS} refers to our baseline which uses self-training to adapt solely the initial layers of the network through pseudo-labels \cite{sheikh2022unsupervised}. \textbf{UDAS ConvLoRA (ours)}, for a fair comparison with UDAS we injected ConvLoRA only to the initial layers. \textbf{Our model: ConvLoRA + AdaBN}, builds on the top of UDAS - however, we do not constrain adaptation to initial layers.
Rather we adapt the whole encoder part of the network via ConvLoRA and position the ESH after the encoder. Furthermore, we use AdaBN to integrate target domain running mean and variance for enhanced adaptation.

\subsection{Results and Analysis}
Table \ref{tab_da:performance_stats} shows that our ConvLoRA + AdaBN achieves superior performance over all the other methods with significantly fewer trainable parameters. When compared to the baseline (UDAS~\cite{sheikh2022unsupervised}), our method (UDAS ConvLoRA) outperforms in four out of five target domains. While for Siemens 1.5, our method has a slight decrease in SDS (0.2\% only) compared to UDAS \cite{sheikh2022unsupervised}, it is important to note that our adaptation is achieved with a substantial reduction in trainable parameters, decreasing from 14,160 (UDAS \cite{sheikh2022unsupervised}) to just 3,954 — a remarkable 72.07\% reduction. When we followed our approach (ConvLoRA + AdaBN), our method achieved better accuracy for Siemens 1.5 as well. 

The U-Net architecture we used, has 24.3 million parameters. With our proposed ConvLoRA-based adaptation in the encoder, the trainable parameters were reduced to 57,714— a reduction of 99.80\%.  Moreover, when we use our ConvLoRA adapter in conjunction with AdaBN (abbreviated as UDAS ConvLoRA+AdaBN), it further boosts model adaptation and outperforms all the other methods, without any additional parameters as demonstrated in Table~\ref{tab_da:performance_stats}.  
Hence, both our standalone ConvLoRA adapter and the combination of ConvLoRA and AdaBN are not only parameter-efficient but also yield competitive results when compared to other methods.

\begin{figure}[!t]
\centering
   \includegraphics[width=0.57\columnwidth, height=3.1cm]{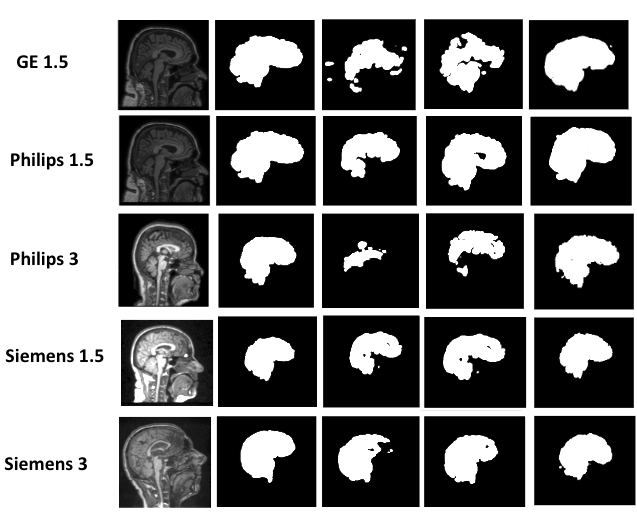}
   \caption{Qualitative Results for target domains of CC359 \cite{souza2018open}. Columns from left to right correspond to input images, ground truth, source U-Net model, UDAS \cite{sheikh2022unsupervised}, and our ConvLoRA + AdaBN. It can be seen that our proposed adaptation has the most visual similarity to the ground truth.}
   \label{da_comparison}
\end{figure}

The qualitative results in Figure~\ref{da_comparison} show that our multi-target UDA method with ConvLoRA + AdaBN (last column) is the most similar to the Ground Truth (second column). We also qualitatively outperform the UDAS work~\cite{sheikh2022unsupervised}. 
\subsection{Ablations}
To identify blocks susceptible to domain shift, we incorporate ConvLoRA adapters into various segments of the network and evaluate their performance, as detailed in Table~\ref{tab:ablations}.
Unlike our baseline (UDAS)~\cite{sheikh2022unsupervised}, we found domain shift is not limited to initial layers. We assessed BN adaptation in the encoder, finding no performance improvement. To evaluate ConvLoRA throughout the network, we employed a siamese network, but using ConvLoRA in the decoder did not enhance performance. Optimal results were achieved by adapting the entire encoder block with ConvLoRA as shown in Table \ref{tab:ablations}. In our adaptation experiments, we test different lengths for the training, ranging from 5 to 20 epochs. The optimal adaptation occurred in just 5 epochs, beyond which overfitting led to decreased performance.
\footnote{This research study was conducted retrospectively using open access CC359 dataset (https://www.ccdataset.com/). Ethical approval was *not* required as confirmed by its license.}
\section{Conclusion}
In this work, we address the problem of unsupervised MT UDA in medical image segmentation with our novel parameter-efficient ConvLoRA adapter, designed specifically for CNNs. We further boost the performance by combining ConvLoRA with AdaBN. We experimentally show that our approach is more accurate and computationally efficient than previous state-of-the-art approaches. We achieve more than 99\% reduction in model parameters while maintaining competitive performance with other UDA segmentation approaches. Our future work is centered on testing the generality of our approach on other medical imaging datasets.


\section{Acknowledgment}
Dedicated to the memory of \textbf{\textit{Dr. Kevin McGuinness}}, whose unwavering support and passion for research will forever guide us.
This research is supported by Science Foundation
Ireland under grant numbers 18/CRT/6183 (ML-LABS
Centre for Research Training).


\bibliographystyle{IEEEbib}
\bibliography{strings,refs}

\end{document}